\documentclass[twocolumn]{bmcart}

\usepackage[utf8]{inputenc} 
\usepackage{soul,color}
  \sethlcolor{green}
  \usepackage{url}
     \usepackage[neveradjust]{paralist}
\usepackage[pdftex]{graphicx}

\startlocaldefs
\endlocaldefs

\begin{document}

%
%
%

\begin{frontmatter}

\begin{fmbox}
\dochead{Ontology}


\title{The African Wildlife Ontology tutorial ontologies: requirements, design, and content}


\author[
   addressref={aff1},                   
   corref={aff1},                       
   email={mkeet@cs.uct.ac.za}   
]{\inits{CM}\fnm{C. Maria} \snm{Keet}}


\address[id=aff1]{
  \orgname{Department of Computer Science, University of Cape Town}, 
  \street{18 University Avenue, Rondebosch},                     %
  \city{Cape Town},                              
  \cny{South Africa}                                    
}





\begin{abstractbox}

\begin{abstract} 
\parttitle{Background} 
Most tutorial ontologies focus on illustrating one aspect of ontology development, notably language features and automated reasoners, but  ignore ontology development factors, such as emergent modelling guidelines and ontological principles. Yet, novices replicate examples from the exercises they carry out. Not providing good examples holistically causes the propagation of sub-optimal ontology development, which may negatively affect the quality of a real domain ontology.

\parttitle{Results} 
We identified 22 requirements that a good tutorial ontology should satisfy regarding subject domain, logics and reasoning, and engineering aspects. 
We developed a set of ontologies about African Wildlife to serve as tutorial ontologies. 
A majority of the requirements have been met with the set of African Wildlife Ontology tutorial ontologies, which are introduced in this paper. The African Wildlife Ontology is mature and has been used yearly in an ontology engineering course or tutorial since 2010 and is included in a recent ontology engineering textbook with relevant examples and exercises. 

\parttitle{Conclusion} 
The African Wildlife Ontology provides a wide range of options concerning examples and exercises for ontology engineering well beyond illustrating only language features and automated reasoning. It assists in demonstrating tasks about ontology quality, such as alignment to a foundational ontology and satisfying competency questions, versioning, and multilingual ontologies. 


\end{abstract}




\end{abstractbox}
\end{fmbox}

\end{frontmatter}



\section{Background}


The amount of educational material to learn about ontologies is increasing gradually, and there is material for different target audiences, including domain experts, applied philosophers, computer scientists, software developers, and practitioners. These materials may include a tutorial ontology to illustrate concepts and principles and may be used for exercises. There are no guidelines as to what such a tutorial ontology should be about and should look like. The two most popular tutorial ontologies are about wine and pizza, which are not ideal introductory subject domains on closer inspection (discussed below), they are limited to the OWL DL ontology language only, and are over 15 years old by now, hence, not taking into consideration the more recent insights in ontology engineering nor the OWL 2 standard with its additional features \cite{OWL2rec}. 

Considering subject domains in the most closely related area, conceptual modelling for relational databa- ses, there is a small set of universes of discourse that are used in teaching throughout the plethora of teaching materials available: the video/DVD/book rentals, employees at a company, a university, and, to a lesser extent, flights and airplanes. 
Neither of these topics for databases lend themselves well for ontologies, for the simple reason that the two have different purposes. It does raise the question as to what would be a suitable subject domain and, more fundamentally, what it is that makes some subject domain suitable but not another, and, underlying that, what the requirements are for an ontology to be a good tutorial ontology.  

In this paper, we will first analyse existing tutorial ontologies and highlight some issues (Section~\ref{sec:othertut}). We then proceed in Section~\ref{sec:reqAndContent} to formulate a preliminary, first, list of requirements that tutorial ontologies should meet, the contents of the African Wildlife Ontology (AWO) tutorial ontologies, and how the AWO fares against the requirements. Further utility is described in Section~\ref{sec:utilityDisc}, as well as a discussion. The scope of this paper is thus to introduce the AWO tutorial ontologies and to frame it in that context. Finally, we conclude the paper in Section~\ref{sec:concl}.

\subsection{Tutorial ontologies: examples of error propagation in teaching}
\label{sec:othertut}

There are two popular tutorial ontologies, being the Wine Ontology and the Pizza ontology, one for being linked to the OWL guide and the other for being linked to the most popular ontology development environment (Prot\'eg\'e). They both have various shortcomings as tutorial ontologies, however, especially concerning modelling practices or styles, which are discussed first.

The Wine ontology\footnote{\url{http://www.w3.org/TR/2003/PR-owl-guide-20031209/wine}} and its related ``Ontology development 101''
\footnote{\url{https://protege.stanford.edu/publications/ontology_development/ontology101-noy-mcguinness.html}} predates OWL 1 with its frames and slots. While the guide contains good suggestions, such as that ``Synonyms for the same concept do not represent different classes'', there are modelling issues, notably that the ontology is replete with the class-as-instance error\footnote{Promoted by the incorrect statement in the guideline ``Individual instances are the most specific concepts represented in a knowledge base.''} (e.g., {\sf TaylorPort} as instance of {\sf Port} and {\sf MalbecGrape} as instance of {\sf Grape} instead of as subclass of), and the sub-optimal object property naming scheme of `hasX' , such as {\sf adjacentRegion} between two {\sf Region}s rather than the reusable and generic {\sf adjacent}. Further, it uses different desiderata in the direct subclassing of wine\footnote{e.g., the likes of {\sf Bordeaux} and {\sf Loire} (region-based) and {\sf Chardonnay} and {\sf Cabernet Sauvignon} (grape-based), and then there are other criteria, like {\sf DessertWine} (food pairing-based grouping) and `wine descriptor' ones ({\sf DryWine}, {\sf RedWine}, {\sf TableWine})}, which does make it interesting for showing classification reasoning (except the undesirable deduction that {\sf DryWine $\equiv$ TableWine}), but is not ideal from a modelling viewpoint. Further, from a tutorial viewpoint: there are many repetitions, such as very many wineries, which distract from the principles, and it lacks annotations. 

The Pizza ontology tutorial was created for the Prot\'eg\'e user manual and OWL DL ontology language \cite{Rector04}.  
It reflects the state of the art at that time, yet much has happened over the past 15 years. For instance, there are new OWL 2 features 
and there are foundational ontologies that provide guidance for representing attributes (cf. Pizza's {\sf ValuePartition}). Pizza's {\sf DomainConcept} throws a learner straight into philosophical debates, which may not be useful to start with, and, for all practical purposes, duplicates {\tt owl:Thing}. Like the  
Wine, 
 it has the `hasX' naming scheme, such as {\sf hasTopping}, including the name of the class it is supposed to relate to, which is a workaround for not having qualified number restrictions (an OWL 1 artefact) and sub-optimal ontological analysis of the relation ({\em in casu}, of how the toppings really relate to the rest of the pizza) that reduces chance of ontology reuse and alignment. Also, this propagates into students' modelling approaches\footnote{such instances were found in ontologies developed by students in earlier instances of the author's course on ontology engineering, such as developing a sandwich ontology with {\sf hasFilling}, an electrical circuit board ontology with {\sf hasIsolator}, furniture with {\sf hasHeadboard} etc.}. Modelling issues are compounded by the ``we generally advise {\em against} doing [domain and range declarations]'' in the tutorial documentation. When one aims to get novices to use Prot\'eg\'e and OWL so as not get too many errors with the automated reasoners, that might make sense, but ontologically, fewer constraints make an ontology less good because it admits more unintended models. Finally, it has repetitive content to show features, which may be distracting, and, as with Wine, there is only one `final' ontology, despite that multiple releases are common in practice. 

Other tutorial ontologies include Family History, University, and Shirt. 
Family History 
\cite{Stevens13} is developed by the same group as Pizza and aims to teach about advanced OWL 2 features and maximise the use of inferencing.  
Loading it in Prot\'eg\'e 5.2 results in three punning errors and trying to classify it returned an OutOfMemoryError (on a MacBookPro, 2.6 GHz and 8GB of memory), which is not ideal to start a tutorial with. Concerning modelling issues, {\sf ParentOfRobert} illustrates one can use individuals in class expressions, but just that the language allows it, does not mean it is ontologically a good idea that must be taught. It also has the `hasX' semantic weakness, very few annotations, {\sf DomainEntity} being subsumed by {\tt owl:Thing}, and multiple data properties. In contrast to Pizza and Wine,   
all the declared instances are instances and the ontology has different versions as one goes along in the chapters. It has some subject domain aspects descending into politics, which would render it unsuitable for teaching in several countries, such as stating that {\sf Sex $\equiv$ Female $\sqcup$ Male} 
(enforcing a gender binary) 
and that {\sf Person $\sqsubseteq$ $\leq 2$ hasParent.Person} (biologically, but not always societally). 

The University ontology also focuses on illustrating OWL features and automated reasoning,   
rather than modelling. 
For instance, it has {\sf AcademicStaff} with sibling {\sf NonAcademicStaff} where a ``non-X'' complement class is sub-optimal, especially when there is a term for it. 
The representation of {\sf Student $\sqsubseteq$ Person} is an advanced modelling aspect that can be improved upon with a separate branch for roles played by an object. The Computer Science Ontology was based on the University Ontology tutorial and contains 
artificial classes, like unions of classes ({\sf ProfessorinHCIorAI}) and underspecified or incorrect individuals like {\sf AI} and {\sf HCI} (e.g., some course instance would be {\sf CS\_AI-sem1-2018}  
instead). 

The Shirt ontology is a tutorial ontology to explain the structure and organisation of the Foundational Model of Anatomy in a simpler way\footnote{\url{http://xiphoid.biostr.washington.edu/fma/shirt_ontology/shirt_ontology_1.php}} and therefore does not have the hasX naming scheme for object properties, it has no data properties and no instances. It has many annotations with explanations of the entities. There are no `interesting' inferences.

Finally, more or less related textbooks were considered 
\cite{Allemang08,Antoniou03,Arp15,Hitzler09,Suarez12}. Only the 
``Semantic Web for the working Ontologist'' (2nd ed.) has sample files for the book's many small examples\footnote{\url{http://www.workingontologist.org/Examples.zip}; Last accessed: 26-11-2018.} with two reoccurring subject domains, being English literature and products.

\subsection{Problems to address}

The previous section described several problems with existing tutorial ontologies. Notably, a recurring shortcoming is that good modelling practices are mostly ignored in favour of demonstrating language features, automated reasoning, and tools. This has a negative effect on learning about ontology development, for tutorial ontology practices are nonetheless seen by students as so-called `model answers' even if it were not intended to have that function.

The ontology survey does not reveal what may be the characteristics of a good tutorial ontology and, to the best of our knowledge, there is no such list of criteria for tutorial ontologies specifically (only for production level domain ontologies,   
such as \cite{DuqueRamos13,KSP15,Poveda12,Schulz12}).  

\subsection{Potential benefits of the African Wildlife Ontology tutorial ontologies}

In order to address these problems, we introduce the African Wildlife Ontology (AWO).
The AWO has been developed and extended over 8 years. It meets a range of different tutorial ontology requirements, notably regarding subject domain, language feature use and automated reasoning, and its link with foundational ontologies on the one hand and engineering on the other. It aims to take a principled approach to tutorial ontology development, which thereby not only may assist a learner, but, moreover from a scientific viewpoint, it might serve as a starting point for tutorial ontology creation or improvement more broadly, and therewith in the future contribute to an experimental analysis of tutorial ontology quality. This could benefit educational material for ontology development.

Also, educationally, there is some benefit to `reusing' the same ontology to illustrate a range of aspects, rather than introducing many small ad hoc examples, for then later in a course, it makes it easier for the learners to see the advances they have made. This is also illustrated with offering multiple versions of the same ontology, which clearly indicate different types of increments.

Finally, the AWO can be used on its own or together with the textbook ``An Introduction to Ontology Engineering'' \cite{Keet18oebook}, which contains examples, tasks and exercises with the AWO.

\section{Construction and content}
\label{sec:reqAndContent}

The construction of the AWO tutorial ontologies has gone through an iterative development process since 2010. This involved various extensions and improvements by design, mainly to address the increasing amount of requirements to meet, and maintenance issues, such as resolving link rot of an imported ontology. Rather than describing the process of the iterative development cycles, we present here a `digest' version of it. First, a set of tutorial ontology requirements are presented together, then a brief overview of the AWO content is described, and subsequently we turn to which of these requirements are met by the AWO. 

\subsection{OE Tutorial ontology requirements}
\label{sec:req}

Tutorials on ontologies may have different foci 
and it is unlikely that an ontology used for a specific tutorial will meet all  
requirements. The ontology should meet the needs for that tutorial or course, and that should be stated clearly. As such, this list is intended to serve as a set of considerations when developing a tutorial ontology. 
Each item easily can take up a paragraph of explanation. We refrain from this by assuming the reader of this paper is sufficiently well-versed in ontology engineering and seeking information on tutorial ontologies. 
For indicative purpose, the requirements are categorised under three dimensions: the subject domain of the ontology, logics \& reasoning, and engineering factors.

\noindent \textit{Subject domain}

\begin{compactenum}[1.]
\item It should be general, common sense, domain knowledge, so as to 
be sufficiently intuitive for non-experts to be able to understand content and add knowledge. Optionally, it may be an enjoyable subject domain to make it look more interesting and, perhaps, also uncontroversial\footnote{Recent general issues include subject domains of exercises that perpetuate stereotypes and simplifications, such as, but not limited to, the gender binary, who can marry whom, and boys with cars.} to increase chance of use across different settings and cultures.\label{t:simple}

\item The  content should 
be not wrong ontologically, neither regarding how things are represented (e.g., no classes as instances)  nor  the subject domain semantics (e.g., whales are mammals, not fish).

\item It needs to be sufficiently international or cross-cultural so that experimentation with a scenario with multiple natural languages for multilingual ontologies is plausible.\label{t:}

\item Its contents should demonstrate diverse aspects succinctly when illustrating a point cf. being repetitive in content.\label{t:succinct}

\item It needs to be sufficiently versatile to illustrate the multiple aspects in ontology development (see below), including the use of core relations such as mereology and meronymy.\label{t:relation}

\item It should permit extension to knowledge that requires features beyond the Description Logics-based OWL species, so as to demonstrate representation limitations and pointers to possible directions of solutions (e.g., fuzzy and temporal aspects, full first order logic).\label{t:ext}

\item The subject domain has to be plausible for a range of use case scenarios (database integration, science, NLP, and so on). \label{t:use}

\end{compactenum}

\noindent \textit{Logics \& Reasoning}

\begin{compactenum}[I.]
\item The ontology should be represented in a logic that has tool support for modelling and automated reasoning.\label{lr:tooling} 

\item The ontology should be represented in a logic that has tool support for `debugging' features that `explain' the deductions such that the tool shows at least 
the subset of axioms involved in a particular 
deduction.\label{lr:}

\item It should permit simple classification examples and easy examples for showing unsatisfiability and inconsistency, meaning as not to involve more than 2-3 axioms in the explanation, and  
also longer  
ones for an intermediate level.\label{lr:simple:class}

\item The standard reasoning tasks should terminate fairly fast ($<5$ seconds) for most basic exercises with the ontology, with the `standard' reasoning tasks being subsumption/classification, satisfiability, consistency, querying and instance retrieval.\label{lr:fast}

\item The representation language should offer some way of importing or linking ontologies into a network of ontologies.\label{lr:import}
 
\item The language should be expressive enough to demonstrate advanced modelling features, such as  irreflexivity and role composition.\label{lr:expressive}

\item The logic should be intuitive for the modelling examples at least at the start; e.g., if there is a need for  ternaries, then that should be used, not a reification or approximation thereof.\label{lr:intuitive}

\end{compactenum}

\noindent \textit{Engineering and development tasks}

\begin{compactenum}[A.]

\item At least some ontology development methods and tools should be able to use the ontology, be used for improvement of the ontology, etc.\label{e:proc}

\item The ontology needs to permit short/simple competency questions (CQs) and may permit long and complicated CQs, which are formulated for the ontology's content and where some can be answered on the ontology and others cannot.\label{e:cq}

\item At least some of the top-level classes in the hierarchy should be straight-forward enough to be easily linked to a leaf category from a foundational ontology (e.g., {\sf Animal} is clearly a physical object, but the ontological status of {\sf Algorithm} is not immediately obvious).\label{e:fo}

\item It should be relatable to, or usable with, or else at least amenable to the use of, ontology design patterns, be they content patterns or other types of patterns, such as the lexico-syntactic or architecture ones.\label{e:odp}

\item It is beneficial if there is at least one ontology sufficiently related to its contents, so that it can be used for tasks such as comparison, alignment, and ontology imports.\label{e:align}

\item It is beneficial if there are relevant related non-ontological resources that could be used for bottom-up ontology development, such as a conceptual model or thesaurus.\label{e:bottomup}

\item It should be able to show ontology quality improvements gradually, stored in different files.\label{e:qual}

\item It should not violate basic ontology design principles (e.g., the data properties issue on hampering reuse).\label{e:designPrinciples}

\end{compactenum}
While this list may turn out not to be exhaustive in the near future, it is expected to be sufficient for introductory levels of ontology development tutorials and courses. 

\subsection{Content of the AWO -- at a glance}

\begin{figure}[h!]
   \centering
   \includegraphics[width=0.47\textwidth]{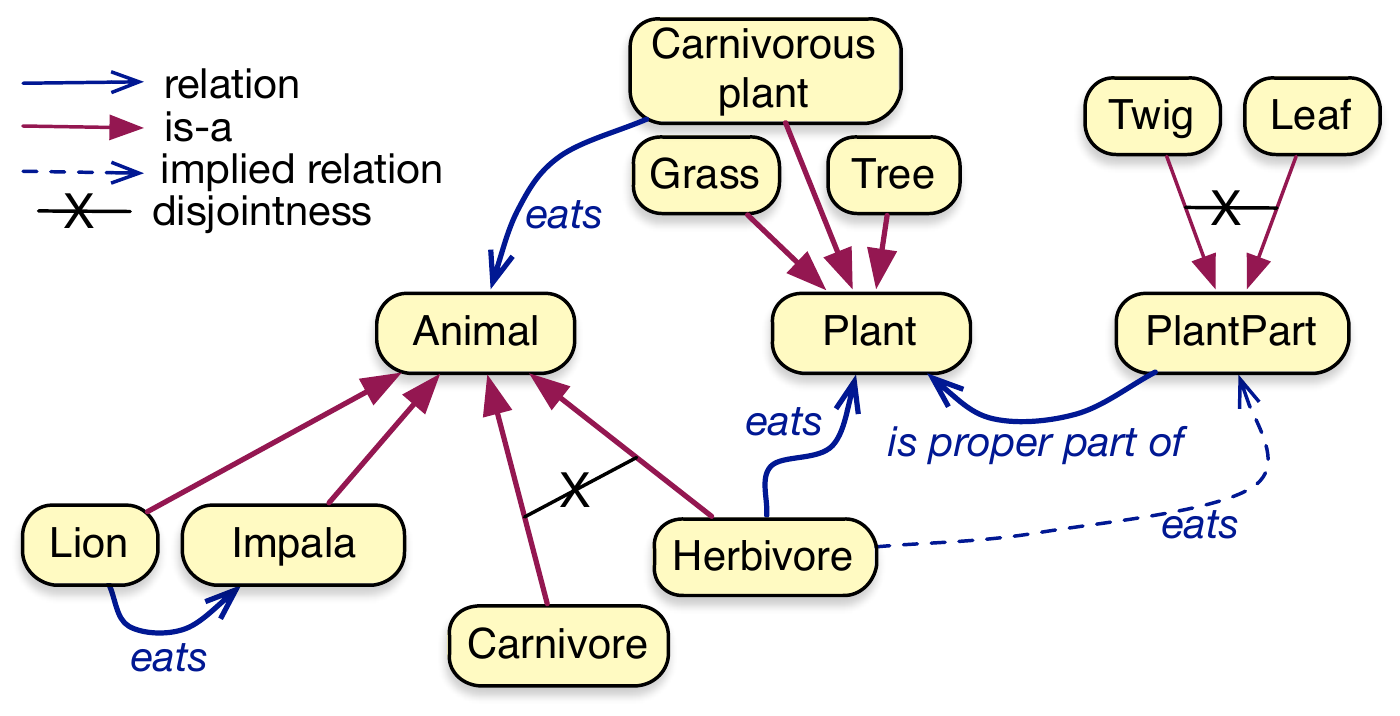} 
  \caption{\csentence{The African Wildlife Ontology at a glance.}
The main classes and relations of the African Wildlife ontology (v1) and an illustrative selection of its subclasses.}\label{fig:AWOoverview}
      \end{figure}

The principal content of the AWO is, in the first stage at least, `intuitive' knowledge about African wildlife.
This subject domain originated from an early Semantic Web book \cite{Antoniou03} (its Section 4.3.1, pp119-133, 1st ed.) that was restructured and extended slightly for its first, basic version; see Table~\ref{tab:awoversions} and Figure~\ref{fig:AWOoverview}. 
 It has descriptions of typical wildlife animals, such as {\sf Lion}s and {\sf Elephant}s, and what they eat, such as {\sf Impala}s (a sort of antelopes) and {\sf Twig}s and (i.e., a logical `or') leaves. Basic extensions in the simple version of the ontology include plant parts, so as to demonstrate parthood and its transitivity, and carnivore vs. herbivore, which make it easy to illustrate disjointness, subsumption reasoning, and unsatisfiable classes, and carnivorous plants to demonstrate logical consequences of declaring domain and range axioms ({\em in casu}, of the {\sf eats} object property). Most elements have been annotated with informal descriptions, and several annotations link to descriptions on Wikipedia. 
 
 Like the aforementioned Family History ontology, there are several versions of the AWO that reflect different stages of learning. In the case of the AWO, this is not specifically with respect to OWL language features, but one of notions of ontology quality and where one is in the learning process. For instance, version 1a contains answers to several competency questions---i.e., quality requirements that an ontology ought to meet \cite{Uschold96}---that were formulated for Exercise 5.1 in the ``Methods and methodologies'' chapter of \cite{Keet18oebook}. Versions 2 and 3, on the other hand, have the AWO aligned to the DOLCE and BFO foundational ontologies, respectively, whose differences and merits are discussed in Chapter 6 of the textbook. Their respective versions with the answers to the related exercises have the name appended with an `a' as well. Version 4 has some contents `cleaned up', partially based on  what the OOPS! tool \cite{Poveda12} detected,  
it uses more advanced language features, and takes steps in the direction of adhering to science (e.g., type of carnivores, distinguishing between types of roots). 
 
There are also four versions in different natural languages, being in isiZulu, Afrikaans, Dutch, and Spanish, which mainly serve the purpose of illustrating issues with multilingual settings of ontology use, which relates to content in Chapter 9 of the textbook.

 \begin{table*}[t]
\caption{AWO ontologies, with their main differences.}
      \begin{tabular}{lp{11.6cm}}
        \hline
         File name  & Difference   \\ \hline
        {\tt AfricanWildlifeOntology.xml}  & This is the file from \url{http://www.iro.umontreal.ca/~lapalme/ift6281/OWL/AfricanWildlifeOntology.xml}, that was based on the description in \cite{Antoniou03}  \\
        {\tt AfricanWildlifeOntologyWeb.owl} & {\tt AfricanWildlifeOntology.xml} + changed the extension to {\tt .owl} and appended the name with Web. This ontology gave at the time (in 2010) a load error in the then current version of Prot\'eg\'e due to the use of {\tt Collection} in the definition of {\sf Herbivore}   \\
        {\tt AfricanWildlifeOntology0.owl} & {\tt AfricanWildlifeOntologyWeb.owl} + that section on {\tt Collection} removed   \\ 
      {\tt AfricanWildlifeOntology1.owl}  & {\tt AfricanWildlifeOntology0.owl} +  several classes and object properties were added (up to $\mathcal{SRI}$ DL expressiveness), more annotations, URI updated (described in  Example 4.1 in \cite{Keet18oebook}) \\ 
      {\tt AfricanWildlifeOntology1a.owl}  & {\tt AfricanWildlifeOntology1.owl} + new content for a selection of the CQs in Exercise 5.1 in \cite{Keet18oebook} (its CQ5, CQ8) and awo\_12 of the CQ dataset \cite{WPLK18})\\ 
      {\tt AfricanWildlifeOntology2.owl}  & {\tt AfricanWildlifeOntology1.owl} + OWL-ised DOLCE ({\tt Dolce-Lite.owl}) was imported and aligned \\ 
      {\tt AfricanWildlifeOntology2a.owl}  & {\tt AfricanWildlifeOntology2.owl} + answers to the questions in Example 6.2 in \cite{Keet18oebook} \\ 
      {\tt AfricanWildlifeOntology3.owl}  & {\tt AfricanWildlifeOntology1.owl} + BFO v1 was imported and aligned \\ 
      {\tt AfricanWildlifeOntology3a.owl}  & {\tt AfricanWildlifeOntology3.owl} + answers to the questions in Example 6.2 in \cite{Keet18oebook} \\                                 
      {\tt AfricanWildlifeOntology4.owl}  & {\tt AfricanWildlifeOntology1.owl} + some things cleaned up (e.g., consistent naming) and added some science content, more OWL language features are used (up to $\mathcal{SRIQ}$), and several educational explanations and questions for further exploration have been added in annotation fields \\
     {\tt AfricanWildlifeOntologyZU.owl} & Mostly {\tt AfricanWildlifeOntology1.owl}  but then in isiZulu, with IRI changed \\
     {\tt AfricanWildlifeOntologyAF.owl} & {\tt AfricanWildlifeOntology1.owl}  but then in Afrikaans, has some IRI issues to resolve \\
     {\tt AfricanWildlifeOntologyNL.owl} & {\tt AfricanWildlifeOntology1.owl} in Dutch, with IRI changed \\
     {\tt AfricanWildlifeOntologyES.owl} & {\tt AfricanWildlifeOntology1.owl} in Spanish, same IRI but different file name \\     
       \hline                                      
      \end{tabular}\label{tab:awoversions}
\end{table*}

\subsection{AWO against the requirements}
\label{sec:awo}
The AWO meets most of the requirements. Concerning the subject domain, the content is general, versatile, not wrong, sufficiently international, and not repetitive (Items~\ref{t:simple}-\ref{t:succinct}). 
The AWO includes the core relation of parthood for, especially, plants and their parts, with optional straightforward extensions with the participation relation (e.g., animals participating in a {\sf Chasing} event) and membership (animal collectives, such as {\sf Herd}; see v4 of the AWO), therewith meeting Item~\ref{t:relation}.
Representation of relevant domain knowledge beyond Description Logics-based OWL species (Item~\ref{t:ext}) could include information about temporal segregation of foraging or commensalism, inclusion of species with distinct successive phases (e.g., {\sf Caterpillar}/{\sf Butterfly}), and the notion of rigidity between what an object is and the role it plays (e.g., {\sf Lion} playing the role of {\sf Predator}; see v4 of the AWO).
Use case scenarios (Item~\ref{t:use}) may be, among others, science of African wildlife, activism on endangered species, and applications such as a database integration and management system for zoos and for tourism websites.

Regarding the logics and reasoning, the AWO is represented in OWL \cite{OWL2rec}, and thus has ample tooling support for knowledge representation, reasoning, and basic debugging/explanation, with ontology development environment tools such as Prot\'eg\'e (Items~\ref{lr:tooling}-\ref{lr:simple:class}). The AWO has both `simple' deductions and more elaborate ones (Item~\ref{lr:simple:class}); e.g., compare {\sf Lion} that is classified as a {\sf Carnivore}, having one explanation involving three axioms, with {\sf Warthog} that is classified as an {\sf Omnivore}, for which there are three justifications computed that each use, on average, five axioms. 
Because the AWO is small, does not make extensive use of individuals and high number restrictions, the reasoner terminates fast under all standard reasoning tasks (Item~\ref{lr:fast}). OWL has the language feature to import other ontologies and it also can be used in other ontology network frameworks, notably the Distributed Ontology, Model, and Specification Language DOL \cite{DOL16} (Item~\ref{lr:import}). While OWL contains expressive features such as role chaining (Item~\ref{lr:expressive}), it, arguably, fails on intuitiveness especially for novices (Item~\ref{lr:intuitive}). Regarding the latter, e.g., novices make errors in the use of existential and universal quantification (for as of yet unclear reasons), which is not known to be a problem when modelling the equivalent in, say, UML Class Diagrams, and there is the elaborate $n$-ary (with $n \geq 3$) approximation issue.

With respect to the engineering aspects, by virtue of the AWO being represented in OWL, there are tools that can process the ontology (Item~\ref{e:proc}), and therewith ontology quality improvement methods can be used with the AWO. They include, e.g., the popular 
Prot\'eg\'e, and various tools for methods and quality, such as test-driven development \cite{KL16} and OOPS! \cite{Poveda12}, and ontology development support activities, such as visualisation and documentation (e.g., \cite{Garijo17,WebVOWL}). There are also a few competency questions that can be answered and that can be easily modelled to be answered, as included in AWO version 1a (Item~\ref{e:cq}), and there are examples and activities to link it to foundational ontologies (AWO versions 2 and 3)  
with easy examples (Item~\ref{e:fo}) (see below, `Utility and Discussion'). 
There are several versions demonstrating various quality improvements (Item~\ref{e:qual}), avoiding violating some basic design principles like data properties and punning hacks (Item~\ref{e:designPrinciples}), and touching upon  some advanced engineering issues with multilingual ontologies (see Table~\ref{tab:awoversions}). 

Where it falls short at the novice level, is an easy way to link it to another ontology (Item~\ref{e:align}) and bottom-up development from non-ontological resources (Item~\ref{e:bottomup}). It is possible and feasible in a mini-project assignment, however; e.g., one could use the freely available wildlife trade data\footnote{\url{https://www.kaggle.com/cites/cites-wildlife-trade-database}} or relate it to the Biodiversity Information Standards\footnote{\url{http://www.tdwg.org/}} for application scenarios, and link it to the Envo Environment ontology \cite{Buttigieg13} or take it easier on the domain knowledge with one of the available tourism ontologies to create an ontology network. 
A bottom-up approach to knowledge acquisition for ontologies is demonstrated with cellfie\footnote{\url{https://github.com/protegeproject/cellfie-plugin}} that implements the M$^2$ DSL \cite{OConnor10} so that a modeller can add content in a spreadsheet and cellfie converts that into axioms in the ontology, as demonstrated in Example 7.1 of the textbook. 
Regarding ODPs (Item~\ref{e:odp}), a content ODP with the current contents is not immediately obvious, but other types of ODPs, such as architectural ones, are easy to illustrate, alike for BioTop \cite{Beisswanger08} but then at the organism-level with an orchestration between foundation, top-domain, and domain-level ontologies.

\section{Utility and discussion}
\label{sec:utilityDisc}

The principal utility of the AWO is to be a concrete machine-processable artefact for the related examples and exercises, which we shall turn to first, and subsequently discuss the tutorial ontology.

\subsection{Use in exercises and examples}

The major utility of the AWO is its use in educational activities for ontology engineering exercises and examples that are described in the ``An Introduction to Ontology Engineering" textbook \cite{Keet18oebook}. It is not intended as a real domain ontology, but it is explicitly designed as a tutorial ontology that has a domain ontology flavour to it. Consequently, the subject domain knowledge about African Wildlife has been kept simple, yet amenable to extensions.

\begin{figure}[h!]
\includegraphics[width=0.47\textwidth]{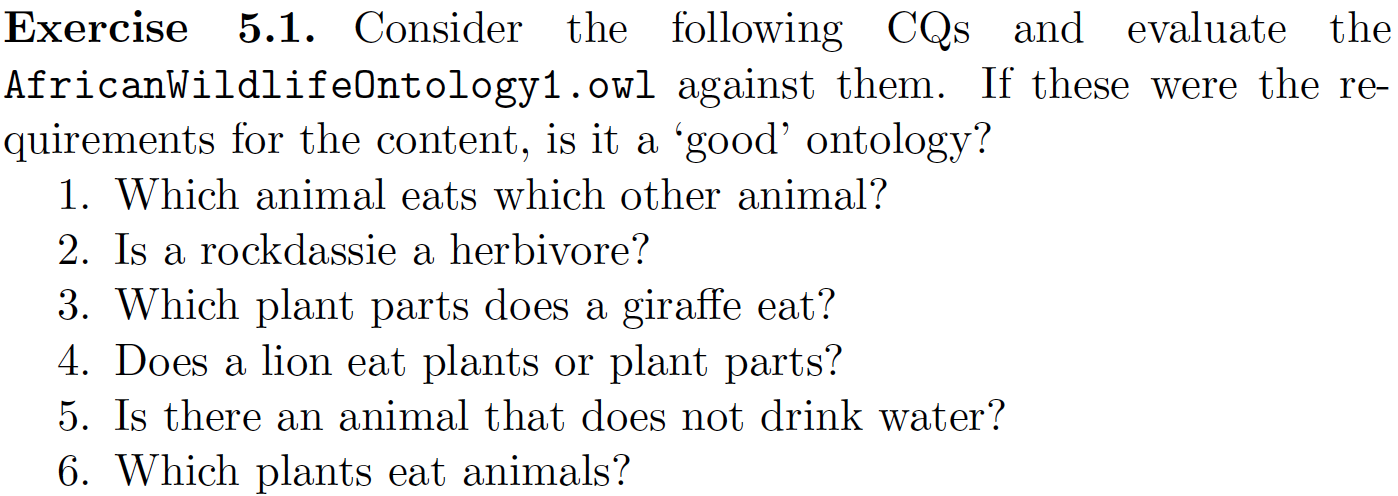}
  \caption{\csentence{Section of an exercise.}
      Screenshot of the first part of Exercise 5.1 in \cite{Keet18oebook}, which lets the student experiment with requirements for the content of an ontology, trying to find that knowledge, and the task of evaluating an ontology on its quality based on its requirements. The high-level notion of a `good' ontology---compared to `less good', `bad', and `worse'---has been introduced earlier in the textbook, which has to be recalled and applied here.}  \label{fig:cq}
      \end{figure}

An example of an exercise is shown in Figure~\ref{fig:cq}, which fits within the broader scope of sensitising the student to the notion of quality of an ontology, with competency questions as one of the options. It also offers a gentle acquaintance with foundational ontologies with some OWL classes that are either easy or fun to categorise or to elicitate lively debate. For instance, impalas die in the process of being eaten by a lion, where both are subclasses of the straightforward {\sf Physical Object} in DOLCE \cite{Masolo03} or {\sf Independent Continuant} in BFO \cite{Arp15}, and {\sf Death} is a type of {\sf Achievement} or {\sf Process boundary}, respectively. The exercises of aligning AWO to DOLCE  is additionally assisted by the D3 decision diagram \cite{KKG13forza}. Death/dying also provides an entry point to the alternate modelling styles of process-as-relation vs. process-as-class representation options. Another core distinction in modelling styles are data properties vs. a hierarchy of qualities, for which a use case of elephant's weight in zoos across the world is used (Section 6.1.1 of \cite{Keet18oebook}).  

While the emphasis in this paper is on modelling and engineering aspects, the AWO is still suitable for teaching about OWL language features and automated reasoning, as noted before regarding the deductions (e.g., ${\sf Lion \sqsubseteq Carnivore}$), and language features use such as transitivity and (ir)reflexivity with parthood. Straightforward examples for demonstrating unsatisfiability are multiple inheritance of {\sf Omnivore} to the disjoint {\sf Carnivore} and {\sf Herbivore} or to set the domain of {\sf eats} to {\sf Animal} resulting in an unsatisfiable {\sf CarnivorousPlant}.

Additional variants of the AWO are in progress, which zoom in on subject domains with corresponding exercises that are not yet covered in the introductory textbook. Among others, a future `version 5' may be the engineering aspects of importing, aligning, and integrating another domain ontology rather than a foundational ontology, such as a module of the Environment Ontology with the habitat information or a tourism ontology, with a corresponding sample answer file. The former option would be more suitable for ontology development in ecology, whereas the latter is a more practical option in a tutorial/course for people in other disciplines. Other themes that have not been covered explicitly yet but easily can be applied to the AWO are modularisation \cite{KK15ao} and Ontology-Based Data Access with its recent tools \cite{Calvanese17}, and it could be assessed against the MIRO guidelines for reporting on ontologies \cite{Matentzoglu18}.

\subsection{Discussion}

The AWO meets most of the tutorial ontology requirements that evolved and extended over the years.  
The AWO goes beyond extant tutorial ontologies that overwhelmingly focus only on demonstrating language features and automated reasoning, or how to use a specific version of a specific tool. In particular, the AWO brings in ontology development aspects, such as competency questions and alignment to a foundational ontology, among others. 

The illustrations of gradual quality improvements---common in ontology development---go beyond the notion that a new version only uses more language features, as in Family History \cite{Stevens13} and  University\footnote{\url{http://owl.man.ac.uk/2005/07/sssw/university.html}}. In particular, there are improvements on aspects regarding, among others, content, naming, annotations, and foundational ontology alignment. 

Also, care has been taken in representing the knowledge, such as avoiding some common pitfalls like the class-as-instance and certain naming issues  like `and', `or' or negation in a term \cite{KSP15}. Unlike other tutorial ontologies, including the popular Pizza and Wine, it is richly annotated with informal descriptions, pointers to introductory domain knowledge, and questions for further exploration of a modelling topic.

Tutorial ontology subject domains such as one's family history, a university, or one's pets are distinctly focussed on individual application scenarios that may serve database development, but do not give an educationally good flavour of typical scopes of domain ontologies. In that regard, pizzas and wines fare somewhat better, which, however, have repetitive content, such as listing all ingredients of pizza topping. Contrast this with animal wildlife, where it suffices already to represent that a lion eats animals to have it classified automatically as a carnivore. The wildlife subject domain is generic rather than specific for one application scenario, and therewith less predisposed to a myopic `my thing only' thinking that is prevalent when students encounter ontologies for a first time that is in line with the UML and EER conceptual data modelling they are familiar with. Last, but not least, African wildlife is obviously relevant for South Africa, where the author and most of her students are based, and it fits with the trend to make curricula regionally relevant.  
This is also reflected in an isiZulu and an Afrikaans version of the ontology and introductory aspects on term use for ontologies in a multilingual setting, as {\sf Impala} and {\sf Rockdassie} are not Standard English words yet they are widely accepted words in South African English.

\section{Conclusions}
\label{sec:concl}

The paper introduced the African Wildlife Ontology tutorial ontologies, which is a set of ontologies used for a variety of ontology development examples and exercises. Considering possible desirable educational outcomes, 22 requirements were formulated that a tutorial ontology should meet. The AWO meets most of these requirements, therewith improving over its predecessors especially reading the notions of evolution of ontology quality several ontology development tasks beyond getting the axioms into an OWL file, such as alignment to a foundational ontology and satisfying competency questions. 

Both the 22 requirements and the AWO are relevant to the field of ontology engineering in particular, especially for enhancing course material, which, it is hoped, will result in further quality improvements of the actual ontologies that developers are building.

\begin{backmatter}


\section*{Competing interests}
  The author declares that she has no competing interests.
  
\section*{Funding}
  The author declares that she has not received project funding for this work.  


\section*{Availability} The AWO is freely available under a CC-BY licence through the textbook's webpage at \url{https://people.cs.uct.ac.za/~mkeet/OEbook/}. 

%

\section*{Acknowledgements}
  The author would like to thank previous ontology engineering course participants on their feedback, which assisted in refining some of the examples and exercises with the AWO.



\newcommand{\BMCxmlcomment}[1]{}

\BMCxmlcomment{

<refgrp>

<bibl id="B1">
  <title><p>{OWL} Pizzas: Practical Experience of Teaching {OWL-DL}: Common
  Errors &amp Common Patterns</p></title>
  <aug>
    <au><snm>Rector</snm><fnm>AL</fnm></au>
    <au><snm>Drummond</snm><fnm>N</fnm></au>
    <au><snm>Horridge</snm><fnm>M</fnm></au>
    <au><snm>Rogers</snm><fnm>L</fnm></au>
    <au><snm>Knublauch</snm><fnm>H</fnm></au>
    <au><snm>Stevens</snm><fnm>R</fnm></au>
    <au><snm>Wang</snm><fnm>H</fnm></au>
    <au><snm>Wroe</snm><fnm>C.</fnm></au>
  </aug>
  <source>Proceedings of the 14th International Conference Knowledge
  Acquisition, Modeling and Management (EKAW'04)</source>
  <publisher>Whittlebury Hall, UK: Springer</publisher>
  <series><title><p>LNCS</p></title></series>
  <pubdate>2004</pubdate>
  <volume>3257</volume>
  <fpage>63</fpage>
  <lpage>81</lpage>
</bibl>

<bibl id="B2">
  <title><p>Manchester Family History Advanced {OWL} Tutorial</p></title>
  <aug>
    <au><snm>Stevens</snm><fnm>R</fnm></au>
    <au><snm>Stevens</snm><fnm>M</fnm></au>
    <au><snm>Matentzoglu</snm><fnm>N</fnm></au>
    <au><snm>Jupp</snm><fnm>S</fnm></au>
  </aug>
  <publisher>UK: University of Manchester</publisher>
  <edition>10</edition>
  <pubdate>2013</pubdate>
  <url>http://owl.cs.manchester.ac.uk/tutorials/fhkbtutorial/</url>
</bibl>

<bibl id="B3">
  <title><p>Semantic Web for the Working Ontologist</p></title>
  <aug>
    <au><snm>Allemang</snm><fnm>D</fnm></au>
    <au><snm>Hendler</snm><fnm>J</fnm></au>
  </aug>
  <publisher>USA: Morgan Kaufmann</publisher>
  <edition>1</edition>
  <pubdate>2008</pubdate>
</bibl>

<bibl id="B4">
  <title><p>A Semantic Web Primer</p></title>
  <aug>
    <au><snm>Antoniou</snm><fnm>G.</fnm></au>
    <au><snm>Harmelen</snm><fnm>F.</fnm></au>
  </aug>
  <publisher>USA: MIT Press</publisher>
  <pubdate>2003</pubdate>
</bibl>

<bibl id="B5">
  <title><p>Building Ontologies with Basic Formal Ontology</p></title>
  <aug>
    <au><snm>Arp</snm><fnm>R</fnm></au>
    <au><snm>Smith</snm><fnm>B</fnm></au>
    <au><snm>Spear</snm><fnm>AD</fnm></au>
  </aug>
  <publisher>USA: The MIT Press</publisher>
  <pubdate>2015</pubdate>
</bibl>

<bibl id="B6">
  <title><p>Foundations of Semantic Web Technologies</p></title>
  <aug>
    <au><snm>Hitzler</snm><fnm>P</fnm></au>
    <au><snm>Krtzsch</snm><fnm>M</fnm></au>
    <au><snm>Rudolph</snm><fnm>S</fnm></au>
  </aug>
  <publisher>USA: Chapman \& Hall/CRC</publisher>
  <edition>1</edition>
  <pubdate>2009</pubdate>
</bibl>

<bibl id="B7">
  <title><p>Ontology Engineering in a Networked World</p></title>
  <publisher>Germany: Springer</publisher>
  <editor>Mari Carmen Su\'arez-Figueroa and Asunci\'on G\'omez-P\'erez and
  Enrico Motta and Aldo Gangemi</editor>
  <pubdate>2012</pubdate>
</bibl>

<bibl id="B8">
  <title><p>Evaluation of the OQuaRE framework for ontology quality</p></title>
  <aug>
    <au><snm>Duque Ramos</snm><fnm>A</fnm></au>
    <au><snm>Fern\'andez Breis</snm><fnm>JT</fnm></au>
    <au><snm>Iniesta</snm><fnm>M</fnm></au>
    <au><snm>Dumontier</snm><fnm>M</fnm></au>
    <au><snm>Egana Aranguren</snm><fnm>M</fnm></au>
    <au><snm>Schulz</snm><fnm>S</fnm></au>
    <au><snm>Aussenac Gilles</snm><fnm>N</fnm></au>
    <au><snm>Stevens</snm><fnm>R</fnm></au>
  </aug>
  <source>Expert Systems with Applications</source>
  <pubdate>2013</pubdate>
  <volume>40</volume>
  <issue>7</issue>
  <fpage>2696</fpage>
  <lpage>2703</lpage>
</bibl>

<bibl id="B9">
  <title><p>Pitfalls in Ontologies and TIPS to Prevent Them</p></title>
  <aug>
    <au><snm>Keet</snm><fnm>C. M.</fnm></au>
    <au><snm>Su\'arez Figueroa</snm><fnm>M. C.</fnm></au>
    <au><snm>Poveda Villal\'on</snm><fnm>M.</fnm></au>
  </aug>
  <source>Knowledge Discovery, Knowledge Engineering and Knowledge Management:
  IC3K 2013 Selected papers</source>
  <publisher>Berlin: Springer</publisher>
  <editor>Fred, A. and Dietz, J. L. G. and Liu, K. and Filipe, J.</editor>
  <series><title><p>CCIS</p></title></series>
  <pubdate>2015</pubdate>
  <volume>454</volume>
  <fpage>115</fpage>
  <lpage>131</lpage>
</bibl>

<bibl id="B10">
  <title><p>Validating ontologies with {OOPS!}</p></title>
  <aug>
    <au><snm>Poveda Villal\'on</snm><fnm>M</fnm></au>
    <au><snm>Su\'arez Figueroa</snm><fnm>MC</fnm></au>
    <au><snm>G\'omez P\'erez</snm><fnm>A</fnm></au>
  </aug>
  <source>18th International Conference on Knowledge Engineering and Knowledge
  Management (EKAW'12)</source>
  <publisher>Germany: Springer</publisher>
  <editor>A. ten Teije and others</editor>
  <series><title><p>LNAI</p></title></series>
  <pubdate>2012</pubdate>
  <volume>7603</volume>
  <fpage>267</fpage>
  <lpage>281</lpage>
  <note>Oct 8-12, Galway, Ireland</note>
</bibl>

<bibl id="B11">
  <title><p>Guideline on Developing Good Ontologies in the Biomedical Domain
  with Description Logics</p></title>
  <aug>
    <au><snm>Schulz</snm><fnm>S.</fnm></au>
    <au><snm>Seddig Raufie</snm><fnm>D.</fnm></au>
    <au><snm>Grewe</snm><fnm>N.</fnm></au>
    <au><snm>R\"ohl</snm><fnm>J.</fnm></au>
    <au><snm>Schober</snm><fnm>D.</fnm></au>
    <au><snm>Boeker</snm><fnm>M.</fnm></au>
    <au><snm>Jansen</snm><fnm>L.</fnm></au>
  </aug>
  <source>Technocal Report</source>
  <pubdate>2012</pubdate>
  <url>http://www.purl.org/goodod/guideline</url>
  <note>v1.0</note>
</bibl>

<bibl id="B12">
  <title><p>An introduction to ontology engineering</p></title>
  <aug>
    <au><snm>Keet</snm><fnm>CM</fnm></au>
  </aug>
  <publisher>UK: College Publications</publisher>
  <series><title><p>Computing</p></title></series>
  <pubdate>2018</pubdate>
  <volume>20</volume>
</bibl>

<bibl id="B13">
  <title><p>Ontologies: principles, methods and applications</p></title>
  <aug>
    <au><snm>Uschold</snm><fnm>M.</fnm></au>
    <au><snm>Gruninger</snm><fnm>M.</fnm></au>
  </aug>
  <source>Knowledge Engineering Review</source>
  <pubdate>1996</pubdate>
  <volume>11</volume>
  <issue>2</issue>
  <fpage>93</fpage>
  <lpage>136</lpage>
  <url>https://doi.org/10.1017/S0269888900007797</url>
</bibl>

<bibl id="B14">
  <title><p>Competency Questions and {SPARQL-OWL} Queries Dataset and
  Analysis</p></title>
  <aug>
    <au><snm>Wisniewski</snm><fnm>D</fnm></au>
    <au><snm>Potoniec</snm><fnm>J</fnm></au>
    <au><snm>Lawrynowicz</snm><fnm>A</fnm></au>
    <au><snm>Keet</snm><fnm>CM</fnm></au>
  </aug>
  <source>Technical Report</source>
  <pubdate>2018</pubdate>
  <issue>1811.09529</issue>
  <url>https://arxiv.org/abs/1811.09529</url>
</bibl>

<bibl id="B15">
  <title><p>{OWL} 2 Web Ontology Language Structural Specification and
  Functional-Style Syntax</p></title>
  <aug>
    <au><snm>Motik</snm><fnm>B</fnm></au>
    <au><snm>Patel Schneider</snm><fnm>PF</fnm></au>
    <au><snm>Parsia</snm><fnm>B</fnm></au>
  </aug>
  <source>W3C Recommendation</source>
  <pubdate>2009</pubdate>
  <note>\url{http://www.w3.org/TR/owl2-syntax/}</note>
</bibl>

<bibl id="B16">
  <title><p>Distributed Ontology, Model, and Specification Language</p></title>
  <aug><au><cnm>Object Management Group</cnm></au></aug>
  <pubdate>2016</pubdate>
  <url>http://www.omg.org/spec/DOL/</url>
</bibl>

<bibl id="B17">
  <title><p>Test-Driven Development of Ontologies</p></title>
  <aug>
    <au><snm>Keet</snm><fnm>C. M.</fnm></au>
    <au><snm>Lawrynowicz</snm><fnm>A.</fnm></au>
  </aug>
  <source>Proceedings of the 13th Extended Semantic Web Conference
  (ESWC'16)</source>
  <publisher>Berlin: Springer</publisher>
  <editor>H. Sack and others</editor>
  <series><title><p>LNCS</p></title></series>
  <pubdate>2016</pubdate>
  <volume>9678</volume>
  <fpage>642</fpage>
  <lpage>657</lpage>
  <note>29 May - 2 June, 2016, Crete, Greece</note>
</bibl>

<bibl id="B18">
  <title><p>{WIDOCO:} A Wizard for Documenting Ontologies</p></title>
  <aug>
    <au><snm>Garijo</snm><fnm>D</fnm></au>
  </aug>
  <source>The Semantic Web – ISWC 2017</source>
  <publisher>Berlin: Springer</publisher>
  <editor>d'Amato, C. and others</editor>
  <series><title><p>LNCS</p></title></series>
  <pubdate>2017</pubdate>
  <volume>10588</volume>
  <fpage>94</fpage>
  <lpage>102</lpage>
</bibl>

<bibl id="B19">
  <title><p>{WebVOWL}: Web-based Visualization of Ontologies</p></title>
  <aug>
    <au><snm>Lohmann</snm><fnm>S</fnm></au>
    <au><snm>Link</snm><fnm>V</fnm></au>
    <au><snm>Marbach</snm><fnm>E</fnm></au>
    <au><snm>Negru</snm><fnm>S</fnm></au>
  </aug>
  <source>Proceedings of EKAW 2014 Satellite Events</source>
  <publisher>Berlin: Springer</publisher>
  <series><title><p>LNAI</p></title></series>
  <pubdate>2015</pubdate>
  <volume>8982</volume>
  <fpage>154</fpage>
  <lpage>-158</lpage>
</bibl>

<bibl id="B20">
  <title><p>The environment ontology: contextualising biological and biomedical
  entities</p></title>
  <aug>
    <au><snm>Buttigieg</snm><fnm>PL</fnm></au>
    <au><snm>Morrison</snm><fnm>N</fnm></au>
    <au><snm>Smith</snm><fnm>B</fnm></au>
    <au><snm>Mungall</snm><fnm>CJ</fnm></au>
    <au><snm>Lewis</snm><fnm>SE</fnm></au>
  </aug>
  <source>Journal of Biomedical Semantics</source>
  <pubdate>2013</pubdate>
  <volume>4</volume>
  <fpage>43</fpage>
</bibl>

<bibl id="B21">
  <title><p>Mapping Master: A Flexible Approach for Mapping Spreadsheets to
  {OWL}</p></title>
  <aug>
    <au><snm>O'Connor</snm><fnm>MJ</fnm></au>
    <au><snm>Halaschek Wiener</snm><fnm>C</fnm></au>
    <au><snm>Musen</snm><fnm>MA</fnm></au>
  </aug>
  <source>Proceedings of the International Semantic Web Conference 2010
  (ISWC'10)</source>
  <publisher>Berlin: Springer</publisher>
  <editor>P. F. Patel-Schneider and others</editor>
  <series><title><p>LNCS</p></title></series>
  <pubdate>2010</pubdate>
  <volume>6497</volume>
  <fpage>194</fpage>
  <lpage>208</lpage>
</bibl>

<bibl id="B22">
  <title><p>Bio{T}op: An Upper Domain Ontology for the Life Sciences - A
  Description of its Current Structure, Contents, and Interfaces to {OBO}
  Ontologies</p></title>
  <aug>
    <au><snm>Beisswanger</snm><fnm>E</fnm></au>
    <au><snm>Schulz</snm><fnm>S</fnm></au>
    <au><snm>Stenzhorn</snm><fnm>H</fnm></au>
    <au><snm>Hahn</snm><fnm>U</fnm></au>
  </aug>
  <source>Applied Ontology</source>
  <pubdate>2008</pubdate>
  <volume>3</volume>
  <issue>4</issue>
  <fpage>205</fpage>
  <lpage>212</lpage>
</bibl>

<bibl id="B23">
  <title><p>Ontology Library</p></title>
  <aug>
    <au><snm>Masolo</snm><fnm>C.</fnm></au>
    <au><snm>Borgo</snm><fnm>S.</fnm></au>
    <au><snm>Gangemi</snm><fnm>A.</fnm></au>
    <au><snm>Guarino</snm><fnm>N.</fnm></au>
    <au><snm>Oltramari</snm><fnm>A.</fnm></au>
  </aug>
  <source>WonderWeb Deliverable D18 (ver. 1.0, 31-12-2003).</source>
  <pubdate>2003</pubdate>
  <note>http://wonderweb.semanticweb.org</note>
</bibl>

<bibl id="B24">
  <title><p>Ontology Authoring with {FORZA}</p></title>
  <aug>
    <au><snm>Keet</snm><fnm>CM</fnm></au>
    <au><snm>Khan</snm><fnm>MT</fnm></au>
    <au><snm>Ghidini</snm><fnm>C</fnm></au>
  </aug>
  <source>Proceedings of the 22nd ACM international conference on Conference on
  Information \& Knowledge Management (CIKM'13)</source>
  <publisher>ACM proceedings</publisher>
  <pubdate>2013</pubdate>
  <fpage>569</fpage>
  <lpage>578</lpage>
  <note>Oct. 27 - Nov. 1, 2013, San Francisco, USA.</note>
</bibl>

<bibl id="B25">
  <title><p>An empirically-based framework for ontology
  modularization</p></title>
  <aug>
    <au><snm>Khan</snm><fnm>ZC</fnm></au>
    <au><snm>Keet</snm><fnm>CM</fnm></au>
  </aug>
  <source>Applied Ontology</source>
  <pubdate>2015</pubdate>
  <volume>10</volume>
  <issue>3-4</issue>
  <fpage>171</fpage>
  <lpage>195</lpage>
</bibl>

<bibl id="B26">
  <title><p>Ontop: Answering {SPARQL} queries over relational
  databases</p></title>
  <aug>
    <au><snm>Calvanese</snm><fnm>D</fnm></au>
    <au><snm>Cogrel</snm><fnm>B</fnm></au>
    <au><snm>Komla Ebri</snm><fnm>S</fnm></au>
    <au><snm>Kontchakov</snm><fnm>R</fnm></au>
    <au><snm>Lanti</snm><fnm>D</fnm></au>
    <au><snm>Rezk</snm><fnm>M</fnm></au>
    <au><snm>Rodriguez Muro</snm><fnm>M</fnm></au>
    <au><snm>Xiao</snm><fnm>G</fnm></au>
  </aug>
  <source>Semantic Web Journal</source>
  <pubdate>2017</pubdate>
  <volume>8</volume>
  <issue>3</issue>
  <fpage>471</fpage>
  <lpage>487</lpage>
</bibl>

<bibl id="B27">
  <title><p>{MIRO}: guidelines for minimum information for the reporting of an
  ontology</p></title>
  <aug>
    <au><snm>Matentzoglu</snm><fnm>N</fnm></au>
    <au><snm>Malone</snm><fnm>J</fnm></au>
    <au><snm>Mungall</snm><fnm>C</fnm></au>
    <au><snm>Stevens</snm><fnm>R</fnm></au>
  </aug>
  <source>Journal of Biomedical Semantics</source>
  <pubdate>2018</pubdate>
  <volume>9</volume>
  <fpage>6</fpage>
</bibl>

</refgrp>
} 

\end{backmatter}
\end{document}